\documentclass[10pt,twocolumn,letterpaper]{article}

\usepackage{wacv}              %

\usepackage{graphicx}
\usepackage{amssymb}
\usepackage{booktabs}
\usepackage{amsmath}

\DeclareMathOperator*{\argmax}{arg\,max}

\usepackage{adjustbox}
\usepackage{graphicx}
\usepackage{multirow}
\usepackage{xcolor}
\usepackage{makecell}
\usepackage{pifont}
\usepackage{import}
\usepackage{color}

\usepackage[pagebackref]{hyperref}
\usepackage[utf8]{inputenc}

\usepackage{cuted}
\usepackage{capt-of}
\usepackage[toc,page]{appendix}

\definecolor{tabfirst}{rgb}{0.7, 1.0, 0.7} %
\definecolor{tabsecond}{rgb}{1, 1, 0.7} %
\definecolor{tabthird}{rgb}{1, 0.85, 0.7} %
\newcommand{\colorboxbold}[2]{%
  \colorbox{#1}{\textbf{#2}}%
}

\newcommand{\ours}{\text{LBG}}

\usepackage[capitalize]{cleveref}
\crefname{section}{Sec.}{Secs.}
\Crefname{section}{Section}{Sections}
\Crefname{table}{Table}{Tables}
\crefname{table}{Tab.}{Tabs.}

\def\wacvPaperID{2754} %
\def\confName{WACV}
\def\confYear{2025}

\title{Lifting by Gaussians: A Simple, Fast and Flexible Method for 3D Instance Segmentation}

\author{Rohan Chacko, Nicolai H{\"a}ni, Eldar Khaliullin, Douglas Lee, Lin Sun\\
Magic Leap Inc.\\
}

\begin{document}
\maketitle

\begin{strip}\centering
 \includegraphics[width=\textwidth]{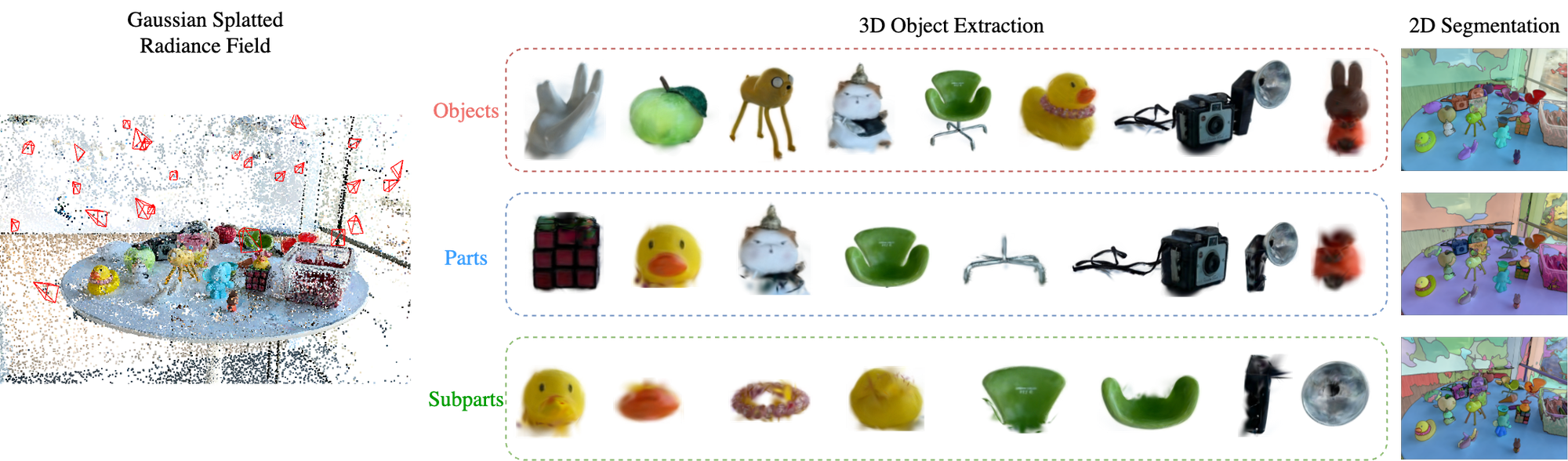}
\captionof{figure}{\textbf{Lifting by Gaussians (LBG).} \ours{} utilizes 2D foundation model masks to segment any pretrained 3DGS field into objects, parts, and subparts without gradient-based learning. For each frame, 2D segmentations are lifted onto the per-pixel max-contributor Gaussian, producing object fragments. These fragments are then merged into coherent, scene-level objects based on both geometric and semantic overlap. Through a hierarchical application of this process, \ours{} extracts high-quality 3D objects, parts, and subparts. In contrast to learning-based methods, LBG achieves this segmentation an order of magnitude faster, enabling new applications like object manipulation in augmented reality.
\label{fig:overview}}
\end{strip}

\begin{abstract}
We introduce Lifting By Gaussians (\ours{}), a novel approach for open-world instance segmentation of 3D Gaussian Splatted Radiance Fields (3DGS). Recently, 3DGS Fields have emerged as a highly efficient and explicit alternative to Neural Field-based methods for high-quality Novel View Synthesis. Our 3D instance segmentation method leverages existing 2D foundation models like SAM, CLIP, and DINO to directly fuse 2D segmentation masks and dense features onto a 3DGS field. Unlike previous approaches, \ours{} requires no per-scene training, allowing it to operate seamlessly on any existing 3DGS reconstruction. Our approach is not only an order of magnitude faster and simpler than existing approaches; it is also highly modular, enabling 3D semantic segmentation of existing 3DGS fields without requiring a specific parametrization of the 3D Gaussians. Furthermore, our technique achieves superior semantic segmentation for 2D semantic novel view synthesis and 3D asset extraction results while maintaining flexibility and efficiency. We further introduce a novel approach to evaluate individually segmented 3D assets from 3D radiance field segmentation methods.
\end{abstract}
    
\section{Introduction}\label{sec:intro}
Semantic scene understanding — segmenting a 3D scene into its constituent objects — is a fundamental challenge in Computer Vision, with wide-ranging applications in Augmented and Virtual Reality (AR/VR), robotics, and autonomous vehicles. In this paper, we introduce a novel 3D scene segmentation method that leverages 2D semantic maps to segment any given 3DGS field~\cite{kerbl_3d_2023}.

2D segmentation has seen rapid advancements, particularly by developing robust 2D foundation models such as the Segment Anything model (SAM)~\cite{kirillov_segment_2023}. However, creating a 3D foundation model that can similarly segment any 3D scene robustly has proven elusive due to the scarcity of annotated 3D data. To bypass the need for annotated 3D data, prior works on 3D segmentation~\cite{shen_distilled_2023,kerr_lerf_2023,qin_langsplat_2024,zhou_feature_2024} instead opt to lift multi-view 2D image segmentation data onto 3D Neural Radiance Fields (NeRF)~\cite{mildenhall_nerf_2021} or 3DGS~\cite{kerbl_3d_2023}. Early techniques focused on a closed set of labels~\cite{genova_learning_2021}, while more recent work has leveraged open-vocabulary models like CLIP~\cite{radford_learning_2021} and DINO features~\cite{caron_emerging_2021} for 2D segmentation. These approaches demonstrate that dense semantic labels optimized via inverse rendering-based formulations can effectively mitigate noisy ground-truth labels. While prior work has achieved great success in embedding 2D semantics onto 3D radiance fields, they either rely on expensive preprocessing steps to enforce multi-view consistency of the semantic images~\cite{ye_gaussian_2023} or suffer from poor quality and long training times due to entanglement of 3D reconstruction and semantic segmentation~\cite{silva_contrastive_2024,ying_omniseg3d_2024,gu_egolifter_2024,kim_garfield_2024}. 

With an ever-growing corpus of existing 3DGS reconstructions, we are interested in quickly segmenting any existing Gaussian Radiance Field into its object, part, and subpart components. Our proposed method LBG accepts two inputs: 1) posed 2D image data, 2) a pre-trained 3DGS field. Using a 2D foundation model, we extract per-image 2D segmentation masks. We then employ a 2D-to-3D lifting approach to assign unique object IDs to Gaussians, creating per-image object fragments in 3D. We then use an incremental merging approach to sequentially merge these object fragments into coherent, scene-level objects. This method enables the segmentation of any existing 3DGS field with significantly reduced processing time compared to contrastive learning methods while delivering higher-quality results as seen in  Table~\ref{tab:training_time}.
 
We validate our approach using standard benchmarks that assess the model's ability to render high-quality semantic maps from novel views. However, similar to~\cite{lee_rethinking_2024}, we argue that the ultimate goal of 3D radiance field segmentation is to generate high-quality 3D assets rather than merely produce compelling 2D segmentation masks. Consequently, we introduce a new evaluation protocol that assesses the rendering quality of individually segmented 3D objects, providing a more accurate measure of 3D segmentation quality. Upon acceptance, we will release the code for our method and evaluation protocol along with our proposed 3D ground-truth datasets.

\begin{table}[!htbp]
\centering
\caption{Training time breakdown (in seconds) of state-of-the-art methods. All timings were benchmarked on a single RTX 3090 GPU. Preprocessing time includes loading the models, extracting 2D masks and computing foundation model features. Our method requires \textbf{10x} less time to achieve similar or higher 3D segmentation quality.}
\label{tab:training_time}
\begin{adjustbox}{width=\linewidth}
\begin{tabular}{lccc}
\toprule
\textbf{Methods} & \textbf{Preprocessing} & \textbf{3D Segmentation} & \textbf{Total}\\
\midrule
    Gaussian Grouping~\cite{ye_gaussian_2023} & 293.44 s & 3629.23 s & 3922.67 s\\
\midrule
    SAGA~\cite{cen_segment_2023} & 1917.66 s& 3289.08 s& 5206.74 s\\
\midrule
    Ours & 422.89 s& 27.07 s& 449.96 s\\
\bottomrule
\end{tabular}
\end{adjustbox}
\end{table}

Our contributions can be summarized as follows:
\begin{itemize}
    \item A training-free 3D instance segmentation approach that utilizes a novel 2D-to-3D lifting strategy to assign Gaussian semantics and an incremental merging procedure based on geometric and semantic overlap criteria. Our approach enables fast 3D segmentation of any existing 3DGS field without needing costly optimization of the 3D semantic field.
    \item A 3D segmentation refinement step that allows extraction of visually appealing 3D assets from existing 3DGS fields.
    \item A novel 3D semantic segmentation dataset and a quantitative evaluation protocol for evaluating the fidelity of individually extracted 3D assets. 
\end{itemize}

\section{Related Work}\label{sec:relwork}
This section reviews key literature on 3D Gaussian Splatting for novel view synthesis and 3D scene segmentation techniques. For a comprehensive survey, we refer readers to~\cite{nguyen_semantically-aware_2024}.

\textbf{Radiance Fields}. Gaussian Splatting (3DGS)\cite{kerbl_3d_2023}, introduced in 2023, has rapidly gained prominence as a real-time method for novel view synthesis, offering superior quality and speed compared to traditional Neural Radiance Fields (NeRFs)\cite{mildenhall_nerf_2021}. Since its inception, 3DGS has spurred research across various domains, including Simultaneous Localization and Mapping (SLAM)\cite{yan_gs-slam_2024,huang_photo-slam_2024,keetha_splatam_2024,matsuki_gaussian_2024}, dynamic scene reconstruction\cite{wu_4d_2024,luiten_dynamic_2024,yang_deformable_2024}, generative 3D/4D content creation~\cite{tang_dreamgaussian_2024,liang_luciddreamer_2024,liu_humangaussian_2024}, and meshing~\cite{guedon_sugar_2024,zhang_rade-gs_2024,huang_2d_2024} among others. This surge of research has led to the development of various parameterizations of 3DGS~\cite{kerbl_3d_2023,huang_2d_2024}. In this work, we introduce versatile semantic segmentation capabilities that can be seamlessly applied to any existing 3DGS field, regardless of the chosen parameterization, enabling semantic scene understanding for the growing corpus of 3DGS fields.

\begin{figure*}[!ht]
    \centering
    \includegraphics[width=\linewidth]{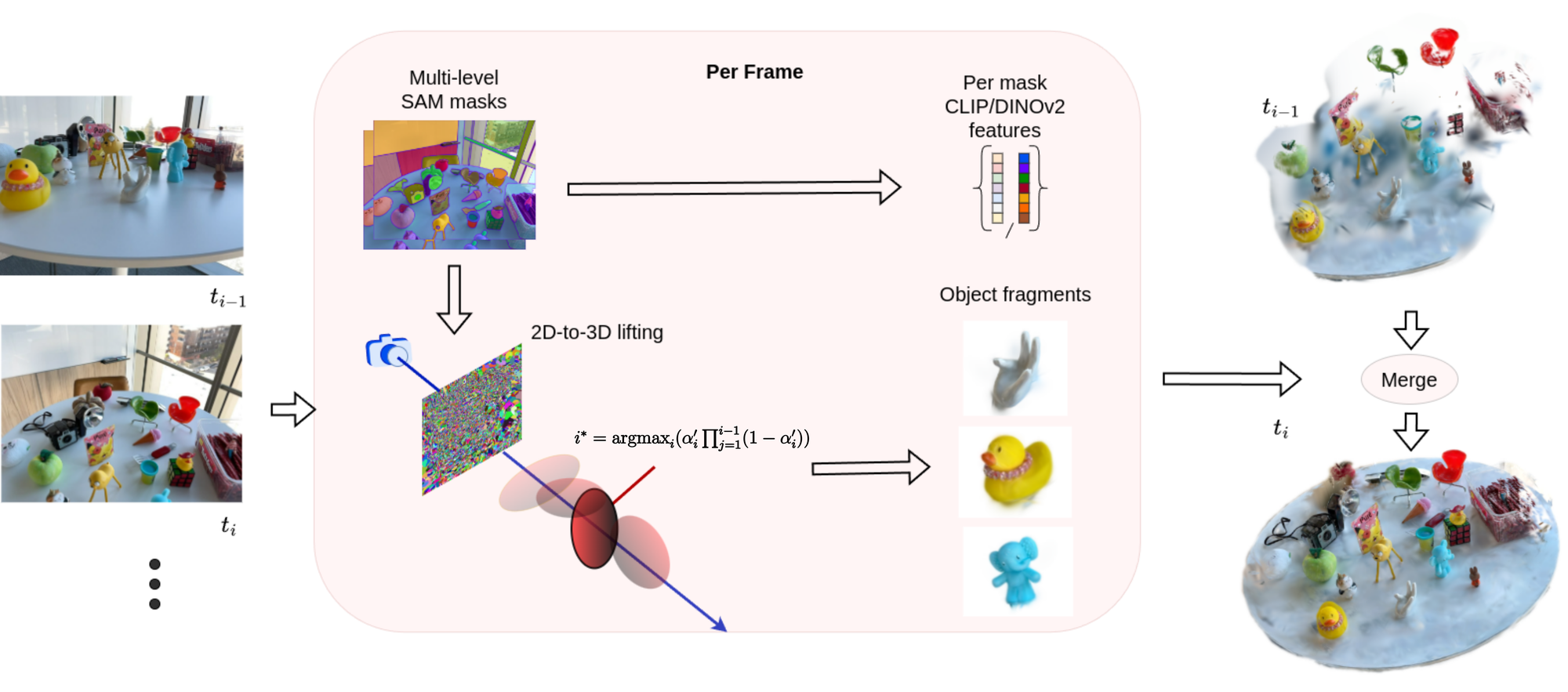}
    \caption{\textbf{\ours{}} constructs an open-vocabulary 3D instance segmentation from a sequence of posed RGB images. A generic 2D instance segmentation model is used to segment \textit{objects}, \textit{parts}, and \textit{subparts} in each RGB image. Semantic feature vectors are extracted for each region, and the masks are lifted to the per-pixel max-contributing Gaussian, generating per-frame 3D object fragments. These fragments are incrementally merged into coherent, scene-level 3D objects. By applying this process hierarchically to the part and subpart masks, \ours{} produces a hierarchical decomposition of any 3DGS scene.}
    \label{fig:segmentation}
\end{figure*}

\textbf{3D Scene Understanding}. Understanding 3D scenes involves inferring the semantic properties of all objects within a scene — a fundamental challenge in 3D computer vision. Early approaches relied heavily on limited 3D ground truth data for tasks like object detection, localization, and segmentation~\cite{dai_scannet_2017,chen_scanrefer_2020,liao_kitti-360_2022}. To circumvent the scarcity of annotated 3D datasets, more recent work has increasingly leveraged 2D supervision~\cite{genova_learning_2021}, which gets fused onto the 3D representation. The advent of large foundation models~\cite{radford_learning_2021,kirillov_segment_2023} has further enabled a shift from closed-set segmentation to an open-world framework, where a broader range of labels can be recognized and utilized~\cite{liang_open-vocabulary_2023,li_language-driven_2022}. These developments in open-world segmentation of neural fields can be broadly categorized into feature distillation and mask-lifting techniques.

\textbf{3D Segmentation through Feature distillation}. Feature distillation~\cite{shen_distilled_2023,zhou_feature_2024} aims to lift 2D features from computer vision foundation models such as CLIP~\cite{radford_learning_2021}, DINO~\cite{caron_emerging_2021}/DINO v2~\cite{oquab_dinov2_2023} or SAM~\cite{kirillov_segment_2023} onto the 3D representation. This is typically achieved through inverse rendering~\cite{shen_distilled_2023,zhou_feature_2024} or feature aggregation~\cite{peng_openscene_2023,jatavallabhula_conceptfusion_2023,gu_conceptgraphs_2024,takmaz_openmask3d_2023}. However, using inverse rendering to learn high-dimensional vectors on each Gaussians in a 3DGS field is computationally expensive, requiring substantial GPU memory and disk storage space, and often results in slower rendering speeds. Langsplat~\cite{qin_langsplat_2024} mitigates some of these issues by compressing latent vectors specific to each scene, albeit at the cost of a lengthy preprocessing step. Moreover, direct aggregation on point clouds lacks the fidelity needed for photorealistic view synthesis. In contrast, our work demonstrates that high-quality 3D segmentation of 3DGS fields can be achieved without learning, offering a faster and more efficient alternative by directly aggregating semantics onto Gaussians.

 \textbf{3D Segmentation through 2D Mask lifting.} Another line of research involves lifting 2D segmentation masks from foundation models like SAM~\cite{kirillov_segment_2023} into 3D space~\cite{ye_gaussian_2023,silva_contrastive_2024,ying_omniseg3d_2024,gu_egolifter_2024,kim_garfield_2024,cen_segment_2023}. A primary challenge in these methods is achieving multi-view consistent segmentation masks. Gaussian Grouping~\cite{ye_gaussian_2023} addresses this issue by using a tracking algorithm, though it struggles with errors under significant viewpoint changes. Other methods rely on contrastive learning to decompose and reconstruct the 3DGS field with semantic information~\cite{silva_contrastive_2024,ying_omniseg3d_2024,gu_egolifter_2024,kim_garfield_2024}. These approaches are data-intensive, computationally expensive, and often entangle the 3D reconstruction with semantic learning. Our approach offers a more efficient alternative by fusing semantics directly onto the Gaussian field, significantly reducing computational costs and memory usage while maintaining or exceeding the segmentation quality of existing methods. 

\section{Method}
We present \textit{Lifting-by-Gaussians} (\ours{}), a novel and efficient method for rapidly lifting 2D semantic information onto any existing 3D Gaussian Splatting reconstruction. Given a set of posed RGB frames, our approach semantically decomposes a 3D environment within minutes. Using a 2D segmentation model, we identify object candidates and associate them across multiple views by applying semantic and geometric similarity measures to the lifted 3D Gaussians. To achieve fine-grained segmentation, we apply this lifting process hierarchically across multiple 2D segmentation scales. Our approach is illustrated in Figure~\ref{fig:segmentation}.

A key advantage of \ours{} is its independence from gradient-based optimization, allowing seamless integration with any scene representation that uses 3D Gaussians, regardless of parameterization (e.g., 3DGS, 2DGS) and without needing to modify the underlying source code. Furthermore, our innovative 2D-to-3D lifting strategy enables direct incorporation of pretrained 2D features, such as those from DINOv2~\cite{oquab_dinov2_2023}, CLIP~\cite{radford_learning_2021}, and SAM~\cite{kirillov_segment_2023}, onto the 3D Gaussians. This bypasses the need for costly training typically required to learn 3D-consistent features, making \ours{} a powerful and flexible solution for 3D scene understanding.

\subsection{3D Gaussian Splatting}
\label{sec:preliminaries}
3D Gaussian Splatting (3DGS)~\cite{kerbl_3d_2023} represents a scene as a set of colored 3D Gaussian primitives. Unlike Neural Radiance Fields (NeRFs)~\cite{mildenhall_nerf_2021}, which have an implicit nature, 3DGS is an explicit representation, where each Gaussian $g_i$ is parameterized by a position $\mu \in \mathbb{R}^3$, scale $S \in \mathbb{R}^3$, rotation $R \in \mathbb{R}^4$, opacity $\alpha \in [0, 1]$ and color $\mathbf{c} \in \mathbb{R}^{3}$ represented as three degrees of spherical harmonics (SH) coefficients. mages are rendered efficiently by splatting these 3D Gaussians onto the image plane using the approach from~\cite{zwicker_ewa_2001}, and the resulting 2D projections are alpha-composited in a depth-first order. The color of a rendered pixel is then computed as:
    
\begin{equation}
c = \sum _{i=1}^{N} c_i \alpha_i' \prod_{j=1}^{i-1}(1 - \alpha_i'), 
\end{equation}
    
where $\alpha_i ' = \alpha_i \cdot \text{e}^{-\frac{1}{2}(x' - \mu ')^T \Sigma'^{-1} (x' - \mu ')}$ defines the contribution of each splatted Gaussian to the pixel. The covariance matrix $\Sigma$ is approximated as $\Sigma = RSS^TR^T$ to ensure positive semi-definiteness during optimization.

\subsection{2D-to-3D Lifting}
\label{sec:gsfusion}
Given a sequence of RGB images $\mathcal{I} = \{I^1, I^2, \dots I^t\}$ and a pretrained 3DGS field $\mathcal{G}$, our goal is to generate a 3D semantic segmentation of the 3DGS field. \ours{} achieves this by incrementally creating a 3D semantic map of the scene. For each frame $I^{t}$ at time, the Gaussian field $\mathcal{G}$ is segmented into a set of objects $\mathcal{O}^{t}$. Each object $o^{t}_{j} = \langle \mathcal{G}_{j}^{t}, f^{t}_{j} \rangle$ is characterized by a set of Gaussians $\mathcal{G}_{j}^{t} \subset \mathcal{G}$ and a semantic feature vector $f_{j}^{t}$. Objects  $o^{t}_{j}$ from every new frame $I^t$ are merged into the existing semantic map $\mathcal{O}^{t-1}$ by either adding to existing objects or instantiating new ones. 
\\\\
\textbf{2D Mask and Feature Extraction:} \ours{} begins by extracting class-agnostic 2D segmentation masks $\{m^{t}_{j}\}_{j=1, ..., M}$ using the Segment Anything (SAM)~\cite{kirillov_segment_2023} model. SAM provides 2D segmentation masks at three semantic levels - \textit{whole}, \textit{part}, and \textit{subpart}. For each extracted mask $m^{t}_{j}$ we also extract semantic features $f^{t}_{j}$ using CLIP~\cite{radford_learning_2021} and DINO~\cite{oquab_dinov2_2023}. Since SAM masks lack inter-frame consistency, a mask fusion strategy is implemented using the learned 3D Gaussians to achieve consistent results across multiple frames.
\\\\
\textbf{Single frame 2D Mask to Gaussian Assignment:} Previous methods~\cite{lyu_gaga_2024} have used pixel-Gaussian associations to lift 2D segmentations into 3D. However, these approaches typically assign object IDs to multiple alpha-blended Gaussians based on a threshold, leading to semantic bleeding artifacts. In contrast, \ours{} follows a pixel-to-Gaussian mapping inspired by~\cite{fang_mini-splatting_2024}, where each pixel is associated with the Gaussian that has the maximum alpha-blending weight. For each pixel $p \in m_{j}^{t}$, we identify the Gaussian $i^* = \argmax_{i} (\alpha_i' \prod_{j=1}^{i-1}(1 - \alpha_i'))$ and assign a unique ID to both the 2D mask and its corresponding 3D Gaussians. This approach creates precise object segments $o^{t}_{j}$ in 3D while minimizing semantic bleed-through.
\\\\
\textbf{Incremental Merging of 3D Object Fragments:}
For each new frame $I^t$, \ours{} merges the detected object fragments $o^{t}_{j} = \langle \mathcal{G}_{j}^{t}, f^{t}_{j} \rangle$ with the existing object map $\mathcal{O}^{t-1}_{j}$, constructed from the previous frames. We first compute the geometric overlap ratio between Gaussians belonging to object fragments from the current and previous frames as $\phi_{geom}(i, j) = \dfrac{|\mathcal{G}^{t}_{i} \cap \mathcal{G}^{t-1}_{j}|}{|\mathcal{G}^{t}_{i}|}$. Specifically, we count the number of overlapping Gaussians in both object maps. We further compute semantic similarity as the normalized cosine similarity between the feature vectors $\phi_{sem}(f^{t}_{i}, f^{t-1}_{j}) = \dfrac{f^{t}_{i} \cdot f^{t-1}_{j}}{2}$. Using these metrics , \ours{} greedily merges new object fragments with existing objects based on the highest similarity scores. If no suitable match is found, a new object is instantiated. Once merged, the Gaussian object segment ids  are updated as $\mathcal{G}^{t}_{i} \cup \mathcal{G}^{t-1}_{j}$, and the semantic feature is updated using a running average: $f_{o_j} = \dfrac{n_{o_j}f^{t-1}_{j} + f^{t}_{j}}{n_{o_j} + 1}$, where $n_{o_j}$ represents the number of fragments associated with object $o_j$ so far.
\\\\
\textbf{Hierarchical Decomposition:} Initially, \ours{} uses SAM’s \textit{object-level} masks to create a high-level scene decomposition. Once the object map $\mathcal{O}_{t}$ is constructed, each object is further split into parts and subparts using incremental merges applied at lower segmentation scales. This hierarchical decomposition is repeated across different levels of granularity, resulting in a scene graph that includes objects, parts, and subparts.
\\\\
\subsection{Post-Processing}\label{sec:post-process}
While our maximum-contributor assignment strategy significantly reduces semantic bleeding, some label noise may persist. To generate clean 3D object assets, we include an optional post-processing step. This involves statistical outlier removal, similar to~\cite{ye_gaussian_2023, cen_segment_2023, ji_segment_2024}, alongside a split-and-merge operation. Under-segmented fragments are split using 3D connected component analysis to identify salient clusters. Unassigned clusters are then merged with salient clusters based on their nearest-neighbor distance and overlap ratio, resulting in refined object segmentations.

\section{Experiments}
\label{sec:results}
This section discusses the experimental setup, including datasets, evaluation metrics, and baselines. We then introduce our novel 3D asset segmentation evaluation protocol and show results on 2D mask rendering. Finally, we analyze our method's design choices to justify our approach.

\subsection{Experimental Setup}
\textbf{Datasets}: We evaluate our method using two distinct datasets. The first is the LERF dataset~\cite{kerr_lerf_2023}, which features in-the-wild scenarios captured with a standard iPhone camera. The second dataset is 3D-OVS~\cite{liu_weakly_2023}, which includes a collection of long-tail object categories. 
For evaluating 2D mask segmentation performance on the 3D-OVS dataset, we adhere to the evaluation protocol outlined in~\cite{liu_weakly_2023}.
For the LERF dataset, we address labeling biases present in earlier annotations~\cite{qin_langsplat_2024, ye_gaussian_2023}, which previously focused on a limited number of central objects per scene. We re-annotate three scenes—\textit{figurines}, \textit{ramen}, and \textit{teatime} — with a denser set of 2D instance labels. We annotate salient objects with 2D instance-level masks and open-world vocabulary annotations for our 2D mask rendering evaluations. Additionally, we use the LERF scenes in our 3D asset segmentation experiments. For these experiments, we first train a high-fidelity 3DGS field for each scene using~\cite{fang_mini-splatting_2024} and then manually clean and refine the 3DGS field by selecting or removing Gaussians to generate a high-quality radiance field for each object.

\textbf{Metrics}: We assess our method's performance on both 2D and 3D semantic segmentation tasks. For novel view synthesis of 2D masks, we use the established evaluation protocols from prior work~\cite{kerr_lerf_2023, ye_gaussian_2023, qin_langsplat_2024, lyu_gaga_2024} and report the mean Intersection over Union (mIoU). 
In our 3D segmentation evaluations, we aim to measure how photorealistically a 3D segmented asset is compared to the underlying 3D model of the object. To evaluate this, we propose to 3D segment the 3DGS scene into individual objects and render 50 images of each object from various angles on a viewing hemisphere on a white background. To asses visual quality we compute perceptual similarity metrics such as peak signal-to-noise ratio (PSNR), structural similarity index (SSIM)\cite{wang_image_2004}, and Learned Perceptual Image Patch Similarity (LPIPS)\cite{zhang_unreasonable_2018} to compare the ground truth and predicted images. 

\textbf{Implementation Details}: Our approach builds upon the Gaussian Splatting (3DGS) framework~\cite{kerbl_3d_2023}. Through our experiments, we identified two key characteristics necessary for accurate 3D asset extraction: minimal noise in the 3DGS field and fewer Gaussians to expedite object segmentation. To address these needs, we adopt Mini-Splatting~\cite{fang_mini-splatting_2024} as our primary 3DGS representation. Mini-Splatting provides a compressed 3DGS model by significantly reducing the number of Gaussians through depth-based re-initialization and re-sampling, achieving a 10x reduction without degrading novel-view synthesis quality. We further enhance Mini-Splatting by incorporating a view consistency score to down-weight Gaussians visible from only a single camera, as these often include artifacts. Visualizations of our improved importance sampling can be found in the appendix.

For our experiments, we train Mini-Splatting~\cite{fang_mini-splatting_2024} 3DGS for 30K iterations. Gaussian Grouping~\cite{ye_gaussian_2023} is trained for 30K iterations across all scenes. For SAGA~\cite{cen_segment_2023}, we test two scenarios: 1) training a standard 3DGS representation for 30K iterations before training the semantic features of SAGA from scratch for 10K iterations and 2) training SAGA on top of our enhanced Mini-Splatting representation for 10K iterations.

Our 2D segmentation model uses SAM with the ViT-H backbone. For CLIP, we utilize the ViT-L/14 variant~\cite{radford_learning_2021}, and for DINOv2, we use the model described in~\cite{fu_featup_2024}.

\begin{table}[!htbp]
\centering
\caption{\textbf{Quanitative results on photorealistic 3D asset segmentation.} \ours{} outperforms previous approaches by better removing spurious Gaussians from the 3D object segmentation.}
\begin{adjustbox}{width=\linewidth}
\begin{tabular}{lcccc}
\toprule
\textbf{Dataset} & \textbf{Methods} & \textbf{PSNR $\uparrow$} & \textbf{SSIM $\uparrow$} & \textbf{LPIPS $\downarrow$}\\
\midrule
\multirow{3}{*}{Figurines} & Gaussian Grouping~\cite{ye_gaussian_2023} & 16.622 & 0.743 & 0.381 \\
                           & SAGA~\cite{cen_segment_2023} & 19.574 & 0.836 & 0.243 \\
                           & Ours & \textbf{28.689} & \textbf{0.934} & \textbf{0.065}\\
\midrule
\multirow{3}{*}{Ramen} & Gaussian Grouping~\cite{ye_gaussian_2023} & 16.586 & 0.732 & 0.367 \\
                           & SAGA~\cite{cen_segment_2023} & 15.570 & 0.693 & 0.444 \\
                           & Ours & \textbf{23.778} & \textbf{0.893} & \textbf{0.013}\\
\midrule
\multirow{3}{*}{Teatime} & Gaussian Grouping~\cite{ye_gaussian_2023} & 17.002 & 0.748 & 0.383 \\
                           & SAGA~\cite{cen_segment_2023} & 16.072 & 0.740 & 0.395 \\
                           & Ours & \textbf{26.020} & \textbf{0.880} & \textbf{0.117}\\
\bottomrule
\end{tabular}
\end{adjustbox}
\label{tab:3d_seg_lerf}
\end{table}

\begin{figure*}[!htbp]
    \centering
    \includegraphics[width=0.86\linewidth]{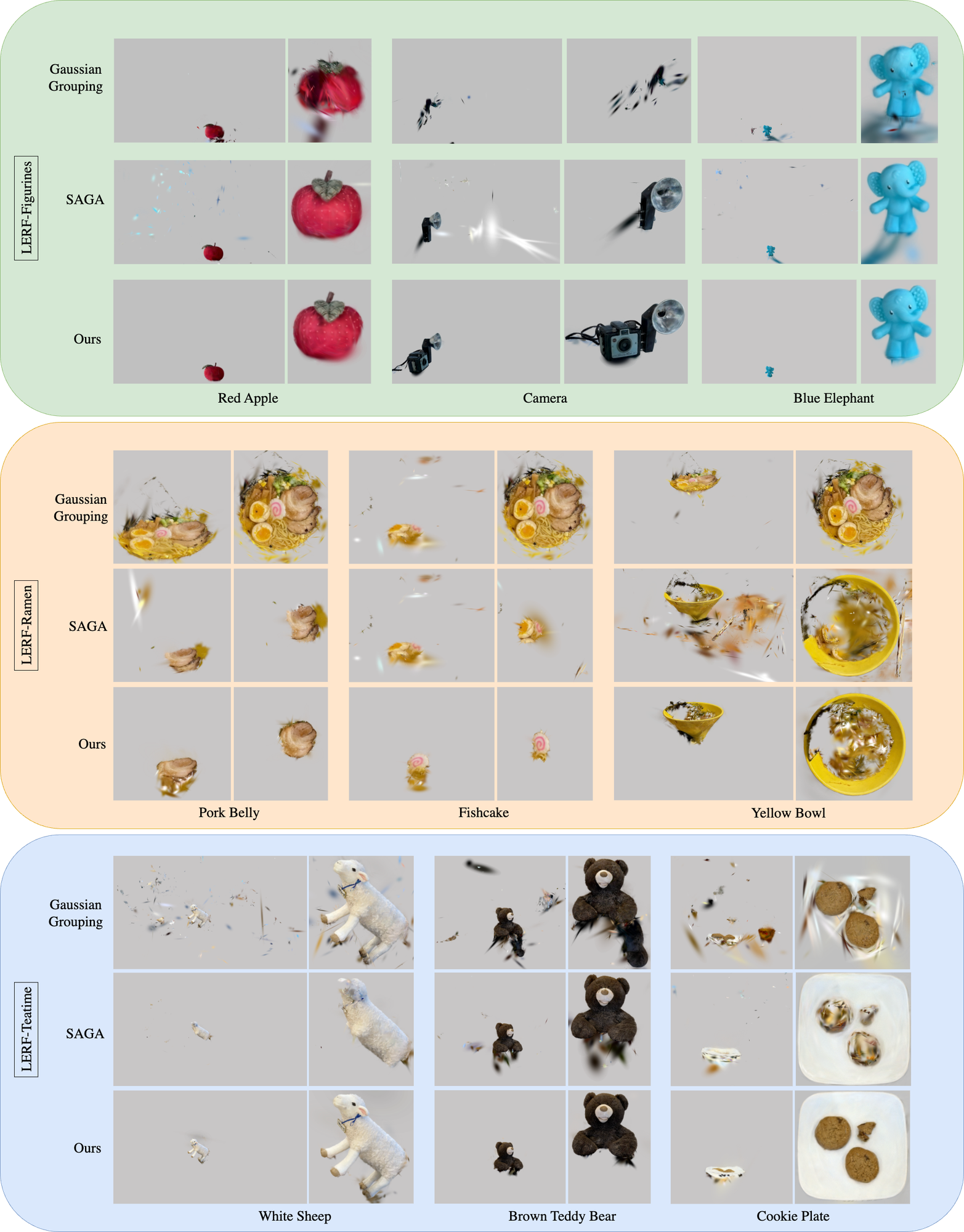}
    \caption{Qualitative comparison on the LERF dataset for 3D Asset extraction. We show three extracted objects per scene, with two different views for each object. Compared to prior methods, the objects extracted from LBG are much cleaner and have fewer noisy artifacts. 3D objects from SAGA and Gaussian Grouping have missing parts and are of lower quality overall.}
    \label{fig:qual_3dseg_lerf}
\end{figure*}

\begin{table*}[!htb]
    \caption{\textbf{2D Novel View Synthesis of 2D Instance Segmentation Masks.} We report mIoU ($\uparrow$) on the LERF dataset~\cite{kerr_lerf_2023} (left) and the 3D-OVS dataset~\cite{liu_weakly_2023}. Best results are highlighted as \colorboxbold{tabfirst}{\textbf{first}}, \colorbox{tabsecond}{second} and \colorbox{tabthird}{third}. In contrast to the baselines, our method was \textbf{not optimized} for this task, but it still performs comparably. Numbers with * are taken from~\cite{cen_segment_2023}}
    \label{tab:2d-seg}
    \begin{minipage}{.5\linewidth}
      \centering
        \begin{adjustbox}{width=0.9\linewidth}
        \begin{tabular}{l c c c}
        \hline
        \textbf{Methods} &  \textit{figurines} & \textit{ramen}  & \textit{teatime}  \\
        \hline
        Gaussian Grouping~\cite{ye_gaussian_2023} & 0.697 & 0.458 & 0.619 \\
        SAGA-MS~\cite{cen_segment_2023} &\colorbox{tabsecond}{0.838} & \colorbox{tabthird}{0.583} & \colorbox{tabthird}{0.673} \\
        SAGA-3DGS~\cite{cen_segment_2023} & \colorboxbold{tabfirst}{0.860} & \colorboxbold{tabfirst}{0.803} & \colorboxbold{tabfirst}{0.874} \\
        \hline
        Ours & \colorbox{tabthird}{0.822} & \colorbox{tabsecond}{0.732} & \colorbox{tabsecond}{0.866} \\
        \hline
        \end{tabular}
        \end{adjustbox}
    \end{minipage}
    \begin{minipage}{.5\linewidth}
      \centering
        \begin{adjustbox}{width=0.9\linewidth}
        \begin{tabular}{l c c c c c c}
        \hline
        \textbf{Methods} & \textbf{Avg.} & \textit{bed} & \textit{bench}  & \textit{room}  & \textit{lawn}  & \textit{sofa} \\
        \hline
        LSEG~\cite{li_language-driven_2022} & 56.0 & 6.0 & 19.2 & 4.5 & 17.5 & 20.6 \\
        OVSeg~\cite{liang_open-vocabulary_2023} & 79.8 & 88.9 & 71.4 & 66.1 & 81.2 & 77.5 \\
        LERF~\cite{kerr_lerf_2023} & 73.5 & 53.2 & 46.6 & 27.0 &  73.7 & 54.8 \\
        3D-OVS~\cite{liu_weakly_2023} & \colorbox{tabthird}{89.5} & \colorbox{tabthird}{89.3} & 92.8 & 74.0 & 88.2 & 86.8 \\
        LangSplat~\cite{qin_langsplat_2024} & 73.02 & 77.8 & 77.3 & 58.4 & 90.9 & 60.2 \\
        Gaussian Grouping~\cite{ye_gaussian_2023} & 88.96 & 64.5 & \colorbox{tabsecond}{95.6} & 
        \colorbox{tabsecond}{96.4} & \colorbox{tabsecond}{97.0} & \colorbox{tabfirst}{91.3}\\
        SAGA~\cite{cen_segment_2023} * & \colorboxbold{tabfirst}{96.0} & \colorbox{tabsecond}{97.4} & \colorbox{tabthird}{95.4} & \colorboxbold{tabfirst}{96.8} & \colorbox{tabthird}{96.6} & \colorboxbold{tabfirst}{93.5} \\
        \hline
        Ours & \colorboxbold{tabsecond}{94.9} & \colorboxbold{tabfirst}{97.7} & \colorboxbold{tabfirst}{96.3} & \colorbox{tabsecond}{95.9} & \colorboxbold{tabfirst}{97.3} & \colorbox{tabthird}{87.4} \\
        \hline
        \end{tabular}
        \end{adjustbox}
    \end{minipage} 
\end{table*}

\subsection{Photorealistic 3D Asset Extraction}
Similar to~\cite{lee_rethinking_2024}, we argue that the ultimate goal of any 3D segmentation method is to extract clean, photorealistic 3D assets. However, the prevailing benchmark metrics in the literature mainly measure the ability of a method to render accurate 2D masks. We argue that rendering 2D masks, especially in the context of 3DGS, can hide small inaccuracies that average out through the alpha-blending process. We, therefore, propose a new evaluation protocol to measure how well any method performs in extracting photorealistic 3D assets. We use a subset of objects from the annotated 3DGS fields for this evaluation. We first match ground-truth and predicted objects by comparing their rendered 2D mask projections across all training views and selecting the one with the smallest average MSE. We extract 3D segmentations for Gaussian Grouping by applying the learned object classifier on each 3D Gaussian. For SAGA, we follow the proposed protocol~\cite{cen_segment_2023} of clustering the 3D feature field using HDBSCAN~\cite{mcinnes_hdbscan_2017}.

As shown in Figure~\ref{fig:qual_3dseg_lerf} and Table~\ref{tab:3d_seg_lerf}, our method outperforms all baselines on this task by a large margin. This can be largely attributed to our simple but effective lifting strategy that avoids bleeding artifacts arising from alpha-blending-based learning on 2D features. Additionally, 
our merge procedure relies on spatial and feature proximity. By only looking at objects accumulated until the previous frame for the merge, we ensure that disconnected Gaussians do not receive spurious labels. Gaussian Grouping suffers from noisy mask predictions and inconsistent segmentation scales. For example, in the \textit{ramen} scene, the method fails to distinguish between objects in the ramen bowl. Similarly in the \textit{figurines} and \textit{teatime} scenes, Gaussian Grouping produces very noisy 3D assets that do not resemble coherent objects. In contrast, SAGA shows much cleaner boundaries in 3D. However, the absence of a well-defined scale hierarchy produces incomplete segmentations like the partial camera segmentation in \textit{figurines}. Consequently, SAGA often tends to over-segment objects into much smaller clusters. Additionally, 3D segmentations extracted from SAGA contain many spurious Gaussians, as can be seen across all three scenes in Figure~\ref{fig:qual_3dseg_lerf}.

\subsection{Novel View Synthesis of 2D Instance Segmentation Masks}
We further evaluate \ours{} on novel view synthesis of 2D instance masks on two datasets, LERF~\cite{kerr_lerf_2023} and 3D-OVS~\cite{liu_weakly_2023}. On the LERF dataset, we compare our method against Gaussian Grouping~\cite{ye_gaussian_2023} and SAGA~\cite{cen_segment_2023}. As each method produces 3D instance segmentation labels differently, we first match ground truth masks to predicted masks by finding the prediction with maximum IoU overlap. To compare fine-grained instance segmentation, we extract multi-level masks where possible. Gaussian Grouping does not provide a hierarchical decomposition of the scene; we only use object-level masks. For SAGA, we compute three segmentation levels, as shown in their paper, namely 0.1, 0.5, and 1.0. For our method, we use the object, part, and subpart hierarchy. 

\begin{figure}[!htbp]
    \centering
    \includegraphics[width=\linewidth]{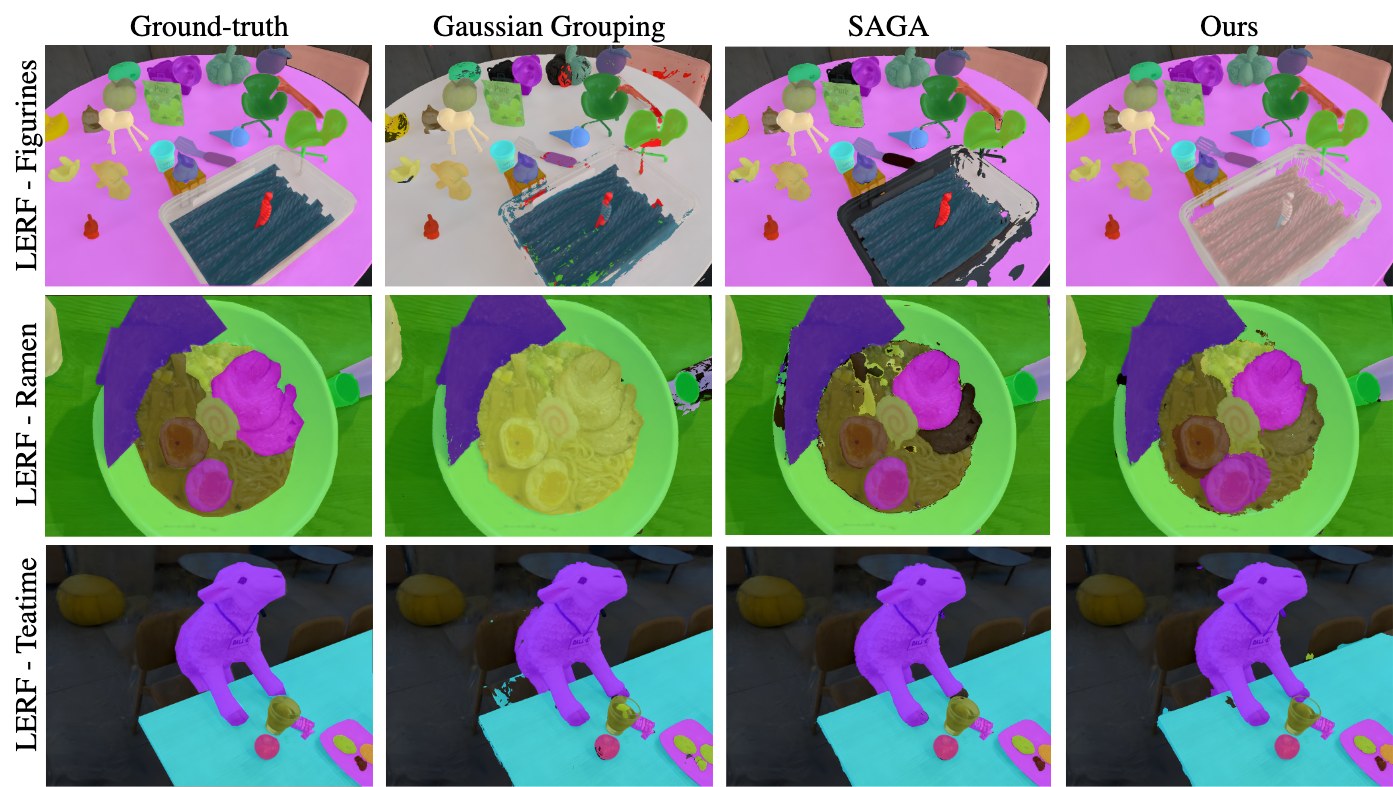}
    \caption{\textbf{Qualitative comparison on novel view synthesis for 2D instance masks}. Black regions are unassigned. We see that our 2D masks are on par with other methods. LBG picks out instances across segmentation scales better than Gaussian Grouping. Compared to SAGA, our method provides more complete masks.}
    \label{fig:2dseg-lerf}
\end{figure}

 In Table~\ref{tab:2d-seg} and Figure~\ref{fig:2dseg-lerf} we show quantitative and qualitative results. Note that while all the baselines are optimized to perform well on the 2D instance mask novel view synthesis task, ours is not. Even so, our method shows competitive performance across the board compared to other approaches (Table~\ref{tab:2d-seg}). As in the 3D segmentation case, Gaussian Grouping can sometimes not distinguish between different object scales, such as in the Ramen bowl in Figure~\ref{fig:2dseg-lerf} (middle). Comparing SAGA and our method, we see that both generate different failure cases. SAGA fails to detect some small objects like the spatula and camera body in \textit{figurines} and part of the pork belly and green onions in \textit{ramen} entirely. Our method generates a complete segmentation map. However, in some cases, \ours{} fails to achieve the desired segmentation granularity. For example, in the \textit{figurines} scene, our method segments the container, Twizzlers, and Waldo together. This can be attributed to inconsistent object segmentation masks from SAM. For additional qualitative results, please see the appendix. 
 
\subsection{Ablations}
\begin{figure}[!htbp]
    \centering
    \begin{minipage}{.5\textwidth}
      \includegraphics[width=\linewidth]{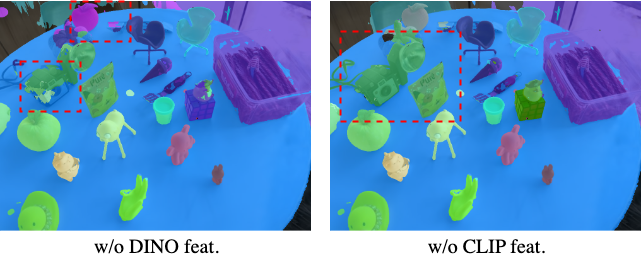}
        \caption{\textbf{Ablation on using CLIP features for merging}. Using only spatial proximity leads to nearby objects being grouped together (red dashed boxes). When using DINO features together with CLIP this error is fixed.}
        \label{fig:ablation_clip_dino}
    \end{minipage}%
\end{figure}

We conduct a series of ablations in Table~\ref{tab:ablation} to evaluate the effect of each design component in LBG. Specifically, we ablate the features of the foundation model used to compute the similarity score in the merging step and the 3DGS reconstruction methods. From Figure~\ref{fig:ablation_clip_dino}, we see that feature selection is critical in merging correct objects. Our post-processing that filters outlier Gaussians provides a slight improvement in segmentation quality. Finally, we evaluate our method with a different choice of a segmentation model~\cite{zhao_fast_2023}).
\\\\
\begin{table}[!htbp]
    \centering
    \caption{\textbf{Ablation experiments.} We show the impact of different design choices on the final 2D instance semgnetation result. We report mIoU ($\uparrow$) on the LERF \textit{figurines} dataset.}
    \begin{adjustbox}{width=0.5\linewidth}
        \begin{tabular}{l c}
        \hline
        \textbf{Ablation} & \textbf{mIoU}\\
        \hline
        Ours, full &  0.822 \\
        w/o filtering & 0.815 \\
        w/o CLIP feat. & 0.782 \\
        w/o minisplatting GS & 0.781 \\
        w/o DINO feat. & 0.779 \\
        w/ Fast-SAM~\cite{zhao_fast_2023} & 0.608 \\
        \hline
        \end{tabular}
        \label{tab:ablation}
    \end{adjustbox}
\end{table} 

\textbf{Effect of post-processing:} Quantitatively, post-processing has only a minimal effect on the 2D instance segmentation results. However, we observe certain cases, where the post-processing step correctly cleans up over-segmented regions and merges them with the correct object. 
\\\\
\textbf{Effect of CLIP and DINO features:} Using just the exact Gaussian overlap as a similarity score tends to aggregate partially overlapping objects as shown in Fig.~\ref{fig:ablation_clip_dino} disregarding their semantic difference. This issue is particularly pronounced for smaller objects, which tend to merge into larger object clusters nearby. This issue is alleviated when considering feature similarity as part of the similarity score. 
\\\\
\textbf{Choice of 3DGS representation:} We observe that the choice of 3DGS representation significantly impacts the final performance. This can be attributed to the post-processing step that scales with the number of Gaussians in the scene. Minisplatting reduces floaters and cloudy artifacts prevalent in vanilla 3DGS reconstructions and has 10x fewer Gaussians. This, in turn, reduces the number of mislabeled Gaussians. Additionally, a smaller GS field allows our algorithm to run much quicker. 
\\\\
\textbf{Choice of segmentation model:} Since our approach is agnostic to the segmentation model used, we apply a faster SAM-variant in our mask extraction step. This speeds up our performance by 4x while providing reasonable 3D segmentations. For more results, please see the appendix.

\section{Conclusion}
\label{sec:conclusion}
In this work, we presented \ours{}, a novel framework for generating high-quality 3D segmentations from any trained 3D Gaussian Splatting (3DGS) field. Our approach uniquely lifts off-the-shelf 2D segmentation masks onto the corresponding per-pixel max-contributor Gaussians, followed by an incremental merging process that consolidates object fragments across frames. By leveraging the max-contributor Gaussians for this 2D-to-3D lifting, \ours{} significantly mitigates semantic bleeding issues that have plagued prior methods, ensuring more accurate and clean object boundaries in the 3D domain.

Experimental results demonstrate that \ours{} not only produces superior 3D assets but also does so at a speed that is an order of magnitude faster than state-of-the-art methods. This substantial improvement in efficiency, combined with high-quality outputs, highlights the transformative potential of \ours{} for large-scale 3D scene segmentation and reconstruction tasks. Moreover, despite not being specifically optimized for novel view synthesis of 2D instance masks, our method delivers competitive performance on 2D instance mask generation, showcasing its versatility across multiple applications.

\subsection{Limitations and Future Work}
While \ours{} achieves efficient, high-quality results, several limitations remain. First, the model loading times hinder real-time applications, which future work could address through optimization or model compression techniques. Second, our method occasionally struggles with segmenting small objects, as the current merging approach may not capture them. A potential solution would involve incorporating a fine-tuning step with minimal training iterations to refine these initial segmentations.

{
    \small
    \bibliographystyle{ieee_fullname}
    \bibliography{wacv}
}

\clearpage

\appendix
\section{Appendix}
In this appendix, we provide further experimental results, including additional 3D segmentation comparisons in Section~\ref{sec:A}, a qualitative comparison with SAGA~\cite{cen_segment_2023} on rendering of 2D masks at novel views through different scales in Section~\ref{sec:B} and qualitative results on the 3D-OVS~\cite{liu_weakly_2023} dataset in Section~\ref{sec:E}. We further show that our method can be used to 3D segment 2DGS fields without modification in Section~\ref{sec:D}. Finally, we show some intuition into our improvements of the Mini-Splatting importance sampling in Section~\ref{sec:F} and conclude by showing an application of our method to lift 2D feature maps, such as DINO, into 3D in Section~\ref{sec:G}.

\section{Additional 3D results}
\label{sec:A}

\begin{figure*}[!htbp]
    \centering
    \includegraphics[width=0.9\linewidth]{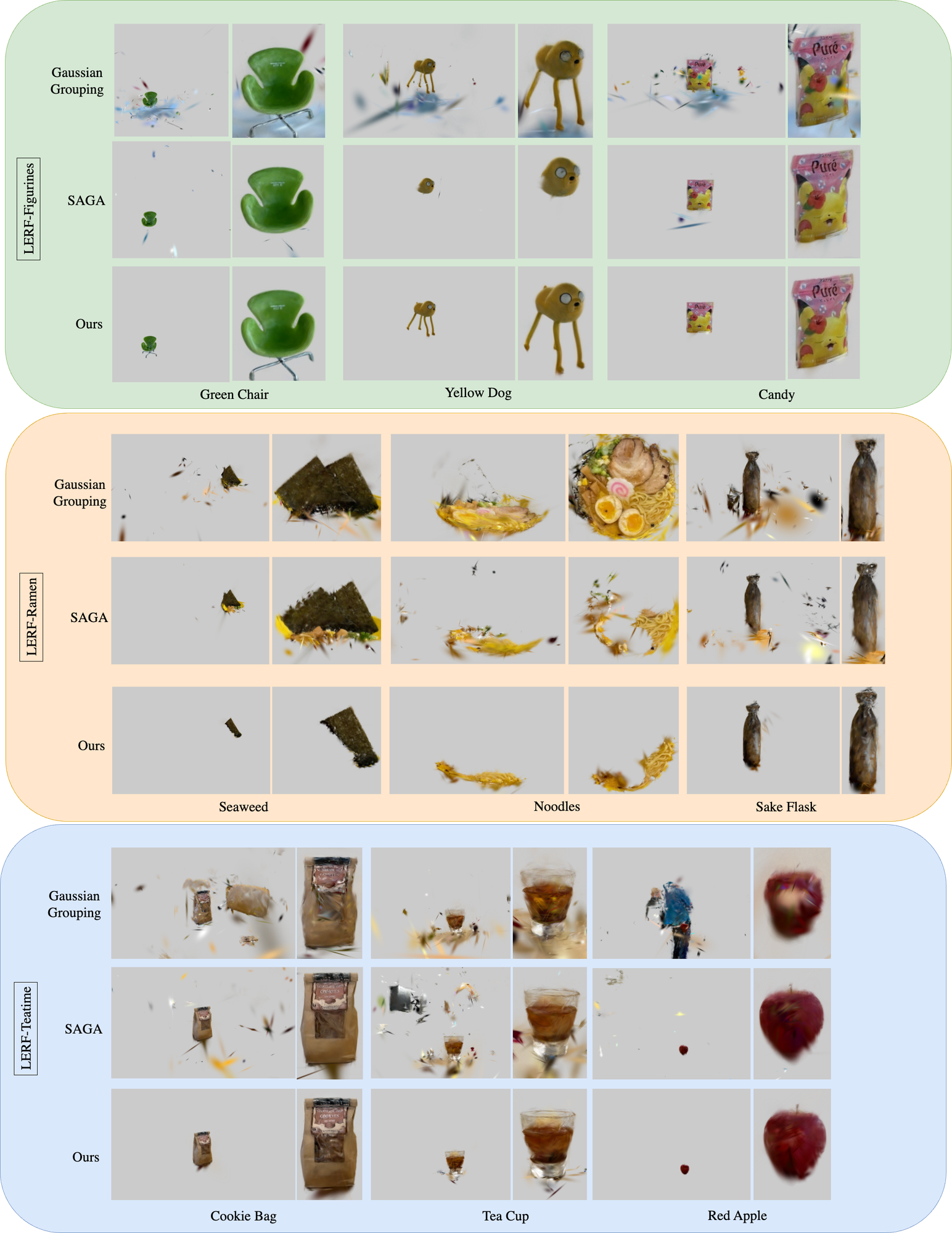}
    \caption{\textbf{Additional 3D segmentation results on LERF dataset}.}
    \label{fig:3d_seg}
\end{figure*}

We show additional results of extracted 3D objects from our LBG method in Figure~\ref{fig:3d_seg}. Contrastive methods like SAGA require a 3D feature clustering step to extract objects, which is prone to floaters and noise. Gaussian Grouping also requires a 3D clustering step which produces noisy 3D objects. Our 3D objects are more coherent and have cleaner boundaries than other methods due to our simple lifting and mask merging strategy. 

\clearpage

\section{Additional 2D results}
\subsection{LERF}\label{sec:B}
\begin{figure}[!htbp]
    \centering
    \includegraphics[width=\linewidth]{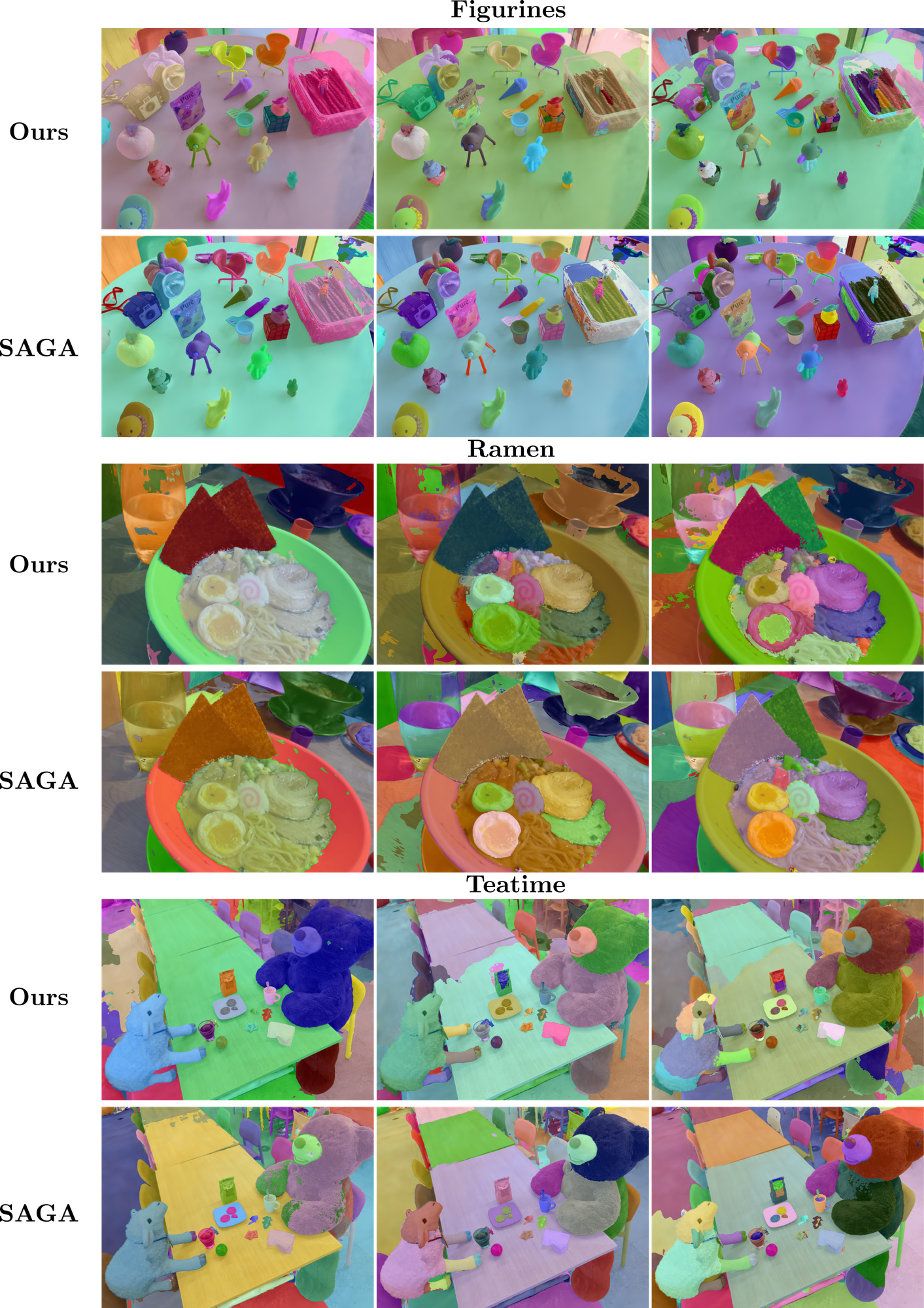}
    \caption{\textbf{Additional results on novel view synthesis for 2D instance masks}. For SAGA, we show images rendered at three levels: 0.1 (left), 0.5 (middle), and 1.0 (right). For our method, we show object level (left), part level (middle), and subpart level (right).}
    \label{fig:2dseg-lerf-appendix}
\end{figure}

We show mask novel view synthesis results on the three LERF scenes in Figure~\ref{fig:2dseg-lerf-appendix}. Specifically, we compare SAGA and \ours{}. For SAGA, we show images rendered at three levels: 0.1 (left), 0.5 (middle), and 1.0 (right). For our method, we show object level (left), part level (middle), and subpart level (right). 

Even though our method is not optimized on the task of novel view synthesis for 2D masks, it performs well, especially on the object level. We can see that SAGA often breaks up objects into parts, even on the top level (camera in figurines, bear in teatime). This is largely due to SAGA using metric diagonal measurements to determine scale without associating these scales back to object/part/subpart decompositions. We argue that instead of using such arbitrary scales it is much more intuitive to break a scene into it's logical parts, starting from complete objects.

\subsection{3DOVS} \label{sec:E}
We compare our method for segmentation against prior methods, such as LSEG~\cite{li_language-driven_2022} and OVSeg~\cite{liang_open-vocabulary_2023}. The numbers for these methods are taken from~\cite{qin_langsplat_2024}. We found that LangSplat evaluations, as described in the paper, led to suboptimal performance due to limited contrast in the learned feature representation. To improve performance, we modified the protocol described in the paper and used a per-scene threshold.  

\begin{figure}[!htbp]
    \centering
    \includegraphics[width=\linewidth]{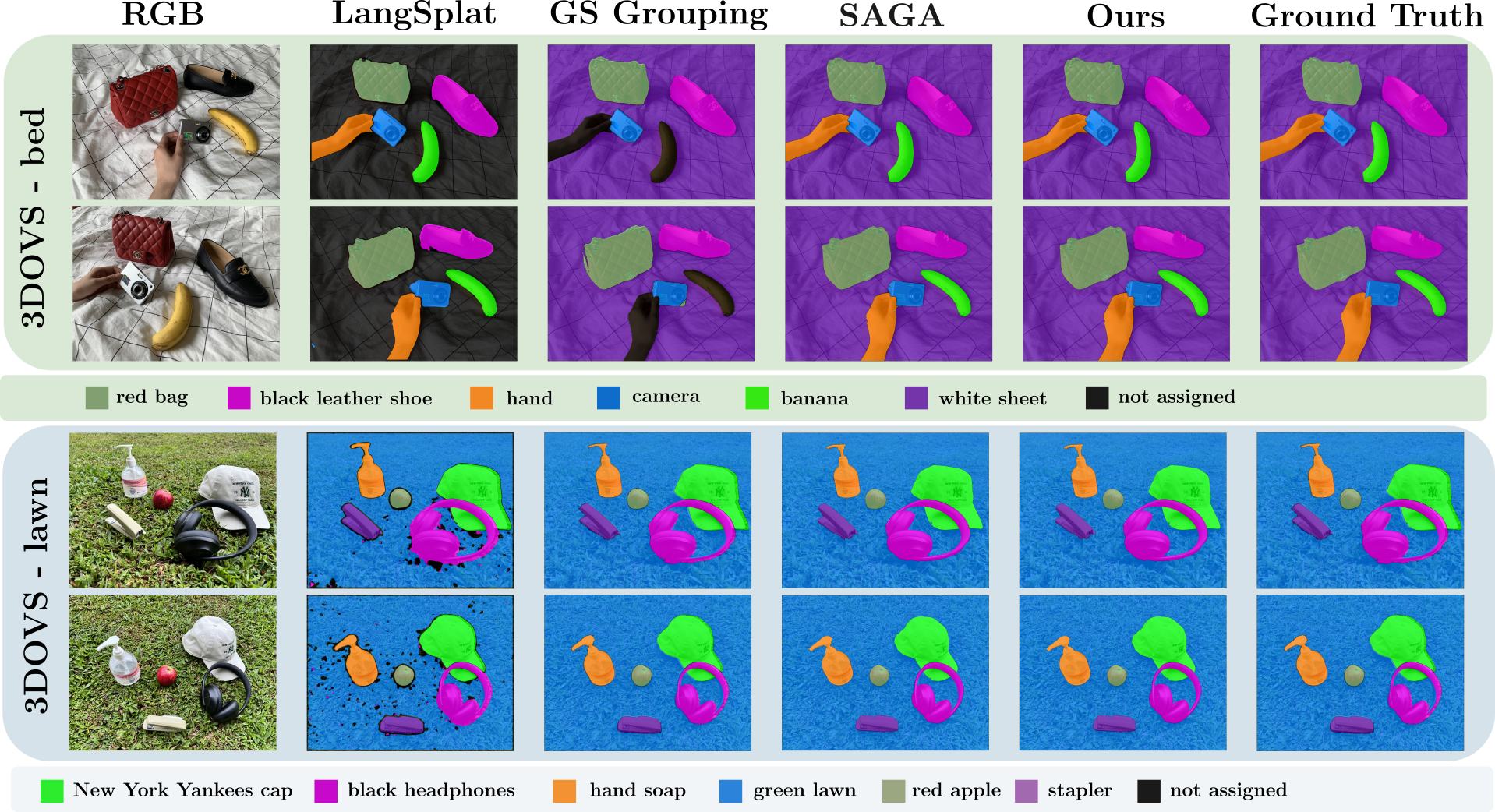}
    \caption{\textbf{Qualitative comparison on the 3DOVS dataset.} Black regions are unassigned. In the bed scene, Gaussian Grouping merges hand and banana objects together, resulting in segmentation failure. Similarly, LangSplat fails to segment the white sheet due to low contrast in the feature space. Our method shows cleaner boundaries compared to both baselines.}
   \label{fig:qual_3dovs}
\end{figure}

On the 3DOVS dataset (Fig.~\ref{fig:qual_3dovs}), our method demonstrates superior performance across the board against most methods and is comparable to SAGA. Notably, LangSplat overlooks the background in the bed scene and exhibits gaps in the lawn scene, attributed to inconsistencies in thresholds and noise within the \textit{subpart} level of the language embeddings. While Gaussian Grouping yields results similar to our method, it often produces less defined boundaries due to tendencies towards over-segmentation. Regions in black are segmentations that were not detected during the evaluation.

\section{Applying our method on 2DGS}\label{sec:D}
\begin{figure}[!htbp]
    \centering
    \includegraphics[width=\linewidth]{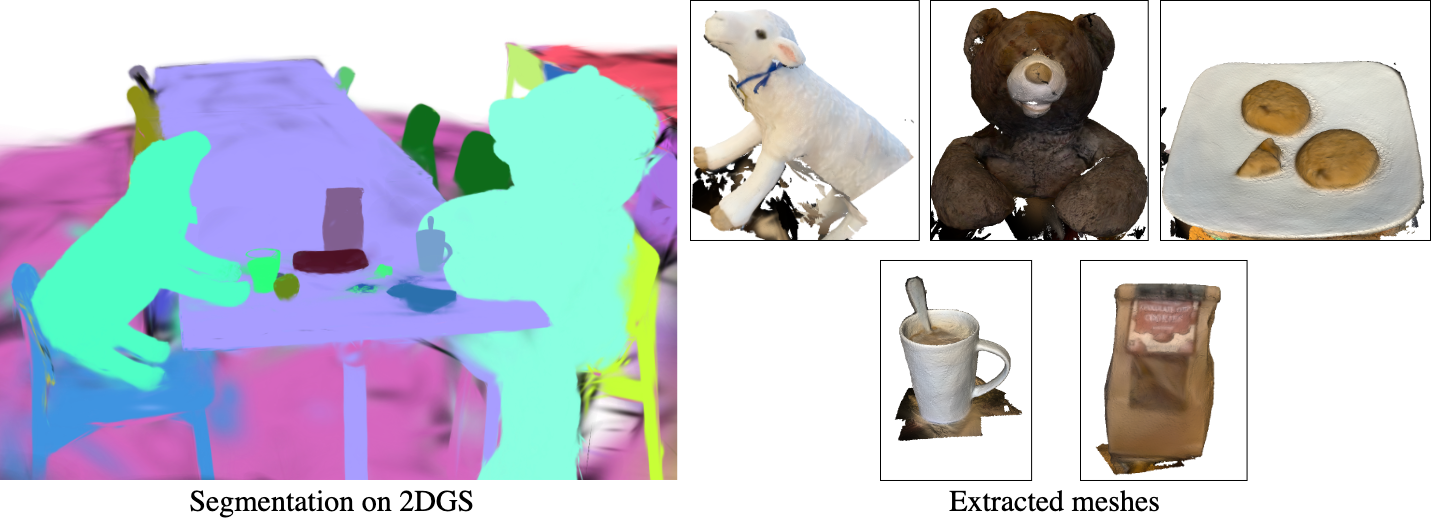}
    \caption{\textbf{LBG Segmentation on 2DGS}. 2DGS with colored Gaussians according to instance IDs (left) and individually extracted meshes (right).}
    \label{fig:2dgs}
\end{figure}

As our method, LBG can consume any Gaussian Splatting-based reconstruction, we apply our method on a scene reconstructed using 2DGS~\cite{huang_2DGS_2024} without any modification. As a consequence of using~\cite{huang_2DGS_2024}, we can produce meshes of the individual segmented objects. Results are shown in Figure~\ref{fig:2dgs}.

\section{Additional Ablations}
\label{sec:C}
\begin{figure}[!htbp]
    \centering
    \includegraphics[width=\linewidth]{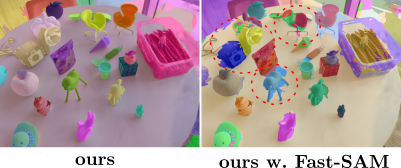}
    \caption{\textbf{Additional ablation results}. We show performance of our method using the standard SAM model (left) and Fast-SAM (right).}
    \label{fig:fastsam}
\end{figure}

We present additional ablation results using our method with Fast-SAM instead of the standard Segment Anything Model for mask extraction. While the Fast-SAM model provides results in near real-time, which is desirable for most applications in robotics and AR, Figure~\ref{fig:fastsam} shows that the results are much worse. We can see that the segmentation lifted with Fast-SAM masks struggles to keep clean object boundaries. Furthermore, even with our post-processing step that merges adjoining clusters, we can see that the Fast-SAM model still contains many objects that cannot be merged using a purely geometric approach.

\section{Improvements to Mini-Splatting} \label{sec:F}
\begin{figure}[!htbp]
    \centering
    \includegraphics[width=\linewidth]{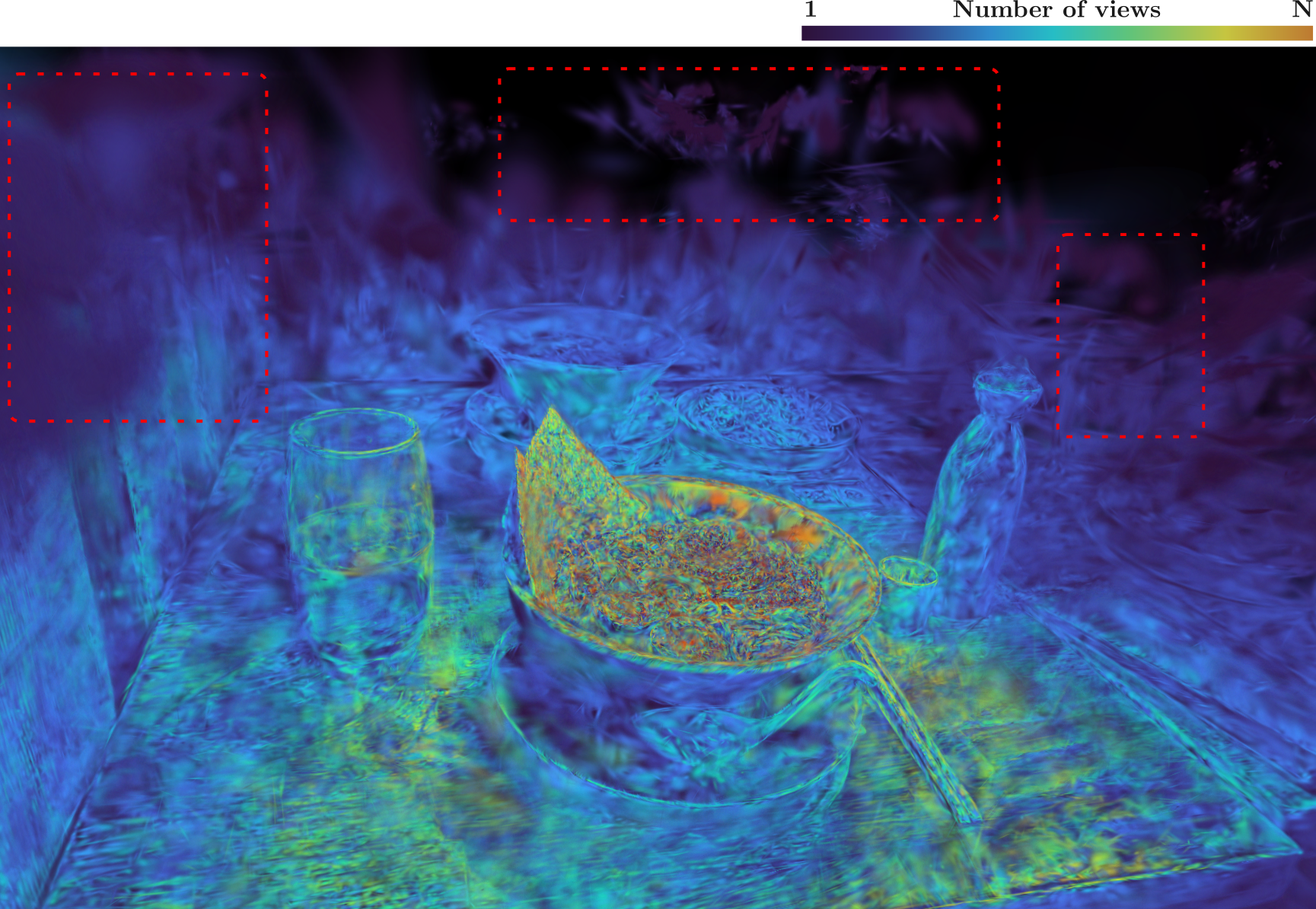}
    \caption{\textbf{Visualization of the number of views which see each Gaussian}. Notice how many structured floaters are only seen by a single view, showcasing visual artifacts from single-view over-fitting.}
    \label{fig:minisplat-vis-score}
\end{figure}

We adopted a technique similar to Mini-splatting to remove floaters from Gaussian Splatting reconstructions. 3D Gaussians are removed based on a probability dictated by an importance score. Initially, we found that using opacity contribution as the basis for this score was insufficient, as it assigned small values to floaters. Many floaters, we discovered, resulted from over-fitting to a single view (see Figure~\ref{fig:minisplat-vis-score}). To address this, we augmented the probability score by considering the number of views a 3D Gaussian maximally contributes to, through a log multiplier on the number of views. This modification, combined with the pruning and resampling strategy from Mini-splatting, effectively reduces floaters, particularly those caused by single-view over-fitting.

\section{DINO Feature lifting}\label{sec:G}

\begin{figure}[!htbp]
    \centering
    \includegraphics[width=\linewidth]{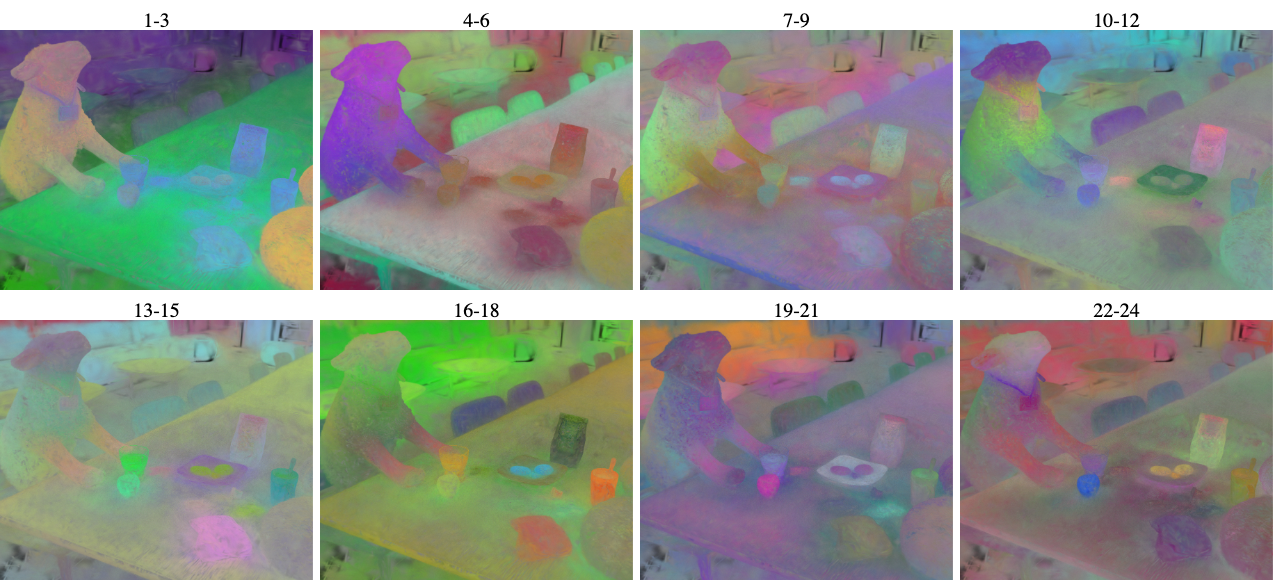}
    \caption{\textbf{Lifting DINOv2 features onto Gaussians}. Using our Lifting-by-Gaussians approach, we lift DINOv2 features and visualize the first 24 PCA components.}
    \label{fig:dino}
\end{figure}

Our approach to lift 2D masks to 3D Gaussian Splatting fields can also lift any 2D foundation model features onto 3D Gaussians using the same strategy. Consequently, we lift DINOv2~\cite{oquab_dinov2_2023} features onto a 3DGS field using our LBG approach. We visualize the first 24 PCA components of the features in Figure~\ref{fig:dino}.

\end{document}